%% file: samplepaper.tex
\documentclass[runningheads]{llncs}
\usepackage[T1]{fontenc}
\usepackage{amsmath}
\usepackage{amssymb}
\usepackage{amsfonts}
\usepackage{mathrsfs}
\usepackage{pifont}
\usepackage{makecell}
\usepackage{booktabs}
\usepackage{array}
\usepackage{multirow}
\usepackage{algorithm}
\usepackage{algpseudocode}
\usepackage{pdflscape}
\usepackage{adjustbox}
\usepackage{makecell}
\usepackage{graphicx}
\usepackage{orcidlink}
\usepackage{float}
\usepackage[section]{placeins}  

\begin{document}
\title{Chameleons do not Forget: Prompt-Based Online Continual Learning for Next Activity Prediction}
\titlerunning{Prompt-Based Online Continual Learning for Next Activity Prediction}

\author{Marwan Hassani\inst{1}\orcidlink{0000-0002-4027-4351} \and
Tamara Verbeek\inst{1} \and
Sjoerd van Straten\inst{1}\orcidlink{0009-0002-5772-9073}}

\authorrunning{M. Hassani et al.}

\institute{Department of Mathematics and Computer Science, Eindhoven University of Technology, The Netherlands\\
\email{\{m.hassani, h.a.j.v.straten\}@tue.nl}}
\maketitle              
\begin{abstract}
Predictive process monitoring (PPM) focuses on predicting future process trajectories, including next activity predictions. This is crucial in dynamic environments where processes change or face uncertainty. However, current frameworks often assume a static environment, overlooking dynamic characteristics and concept drifts. This results in catastrophic forgetting, where training while focusing merely on new data distribution negatively impacts the performance on previously learned data distributions. Continual learning addresses, among others, the challenges related to mitigating catastrophic forgetting. This paper proposes a novel approach called Continual Next Activity Prediction with Prompts (CNAPwP), which adapts the DualPrompt algorithm for next activity prediction to improve accuracy and adaptability while mitigating catastrophic forgetting. We introduce new datasets with recurring concept drifts, alongside a task-specific forgetting metric that measures the prediction accuracy gap between initial occurrence and subsequent task occurrences. Extensive testing on three synthetic and two real-world datasets representing several setups of recurrent drifts shows that CNAPwP achieves SOTA or competitive results compared to five baselines, demonstrating its potential applicability in real-world scenarios. An open-source implementation of our method, together with the datasets and results, is available at: \url{https://github.com/SvStraten/CNAPwP}.

\end{abstract}

\input{introduction}
\input{related}
\input{problem}
\input{main}
\input{experiments}

\section{Conclusion and Future Work} \label{section:conclusion}
In this work, we propose CNAPwP, a novel approach for continual next activity prediction aimed at mitigating catastrophic forgetting. We evaluated its performance using various metrics, including a novel measure called task-specific forgetting, which quantifies the method's memory retention. Experiments conducted on several datasets indicate that CNAPwP achieves the highest average accuracy on two of the four datasets. In one dataset, it ranks second, with Landmark performing the best but being highly inefficient. Therefore, CNAPwP is still regarded as the superior option. Furthermore, analyses of accuracy per event index and task-specific forgetting indicate that our approach exhibits minimal forgetting compared to alternative methods.

Detecting concept drifts was outside the scope of the paper. In the future, it is crucial to explore areas such as integrating a robust concept drift detection mechanism, possibly for both sudden and gradual drifts. Additionally, exploring alternative methods for task comparison might offer sharper, faster, and more efficient ways of performing task similarity detection.

Finally, our experimental results highlighted that the complexity of the event log directly impacts the processing time of the prompting mechanism (cf. Table~\ref{tab:time_per_event}). While CNAPwP maintains robust accuracy in these complex scenarios, the computation cost increases relative to methods such as DynaTrainCDD. We hypothesize that this latency is partly attributed to the current bucketing strategy. Complex logs typically exhibit high variance in trace lengths, which may lead to fragmented batches across numerous buckets, thereby reducing parallelization efficiency. Future research should therefore investigate the specific relationship between log complexity and prompt overhead, with the aim of developing lightweight prompting strategies that optimize this trade-off for highly complex, real-time environments.

\clearpage
\bibliographystyle{splncs04}
\bibliography{sn-bibliography}
\end{document}

%% file: introduction.tex
\vspace{-2mm}
\section{Introduction}
\label{section:introduction}
\vspace{-1mm}
Recent emphasis on predictive process monitoring (PPM) highlights its critical importance by enabling organizations to predict the future trajectories of individual process instances dynamically \cite{ferilli2019activity}, \cite{feldman2013proactive}. PPM offers decision support that empowers organizations to make adjustments to their processes. Next activity prediction holds significant relevance due to its ability to anticipate and plan for future actions within business processes. This predictive capability enables proactive management of resources, ensuring they are deployed effectively to meet upcoming demands. For example, accurate next activity prediction in customer service enables organizations to plan ahead, optimize resources, and reduce response times, leading to more efficient operations and improved customer satisfaction \cite{wolters2022predicting}.

Offline PPM involves two main phases: training and testing. Performing next activity prediction in an online setting offers distinct advantages over offline methods. In an online environment, predictive models continuously learn and adapt in real time as new data streams in, more accurately reflecting the current state of operations. This dynamic approach allows organizations to respond swiftly to changing conditions and optimize processes continuously, whereas offline methods typically rely on static datasets that might be limited in capturing evolving patterns and trends effectively \cite{wolters2022predicting}, \cite{kosciuszekonline}, \cite{guzman2022log}.


A significant challenge in online next activity prediction arises from adapting to new concepts following shifts in data distribution. While real-time model updates enable responsiveness to evolving trends and operational changes, they also introduce the risk of \textit{catastrophic forgetting} \cite{kirkpatrick2017overcoming}. This phenomenon occurs when the prediction model overwrites previously learned information with a new data distribution, potentially leading to a loss of accuracy or reliability in predicting next activities when refacing previously learned data distributions. Balancing dynamic model updates with the retention of critical historical knowledge is crucial, especially given constraints such as storage limitations and privacy concerns that prevent organizations from retaining all historical data indefinitely.
Developing robust online prediction models requires efficient use of historical data while integrating new information. In process mining, recurring drifts from factors such as seasonal changes or shifts in customer behavior make it essential to prevent catastrophic forgetting. Neglecting recurring tasks would compromise the model's long-term performance.

Only recently algorithms have been developed for online PPM \cite{hurtado2023continual} and specifically for online next activity prediction \cite{wolters2022predicting}, \cite{kosciuszekonline}. In contrast, numerous algorithms already exist for online image classification that avoid catastrophic forgetting, including \cite{wang2022dualprompt}, \cite{L2P}, \cite{rolnick2019experience}, \cite{kirkpatrick2017overcoming}. By leveraging the techniques utilized in such algorithms, there is potential to create high-performance online next activity prediction models for this purpose. Unlike independent data points in classification, next activity prediction requires capturing temporal dependencies, which presents unique challenges.

\begin{figure}
    \centering
    \includegraphics[width=1\linewidth]{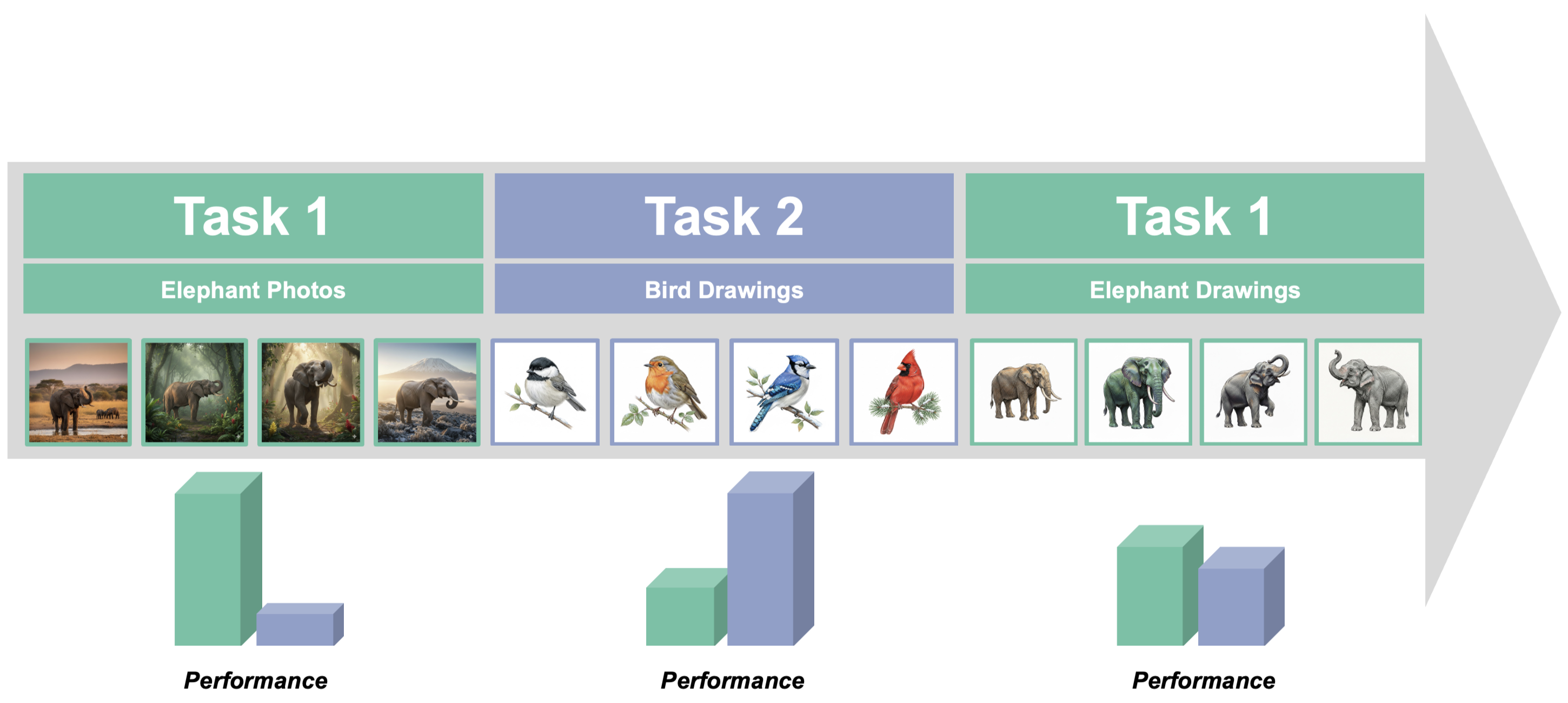}
    \caption{Catastrophic forgetting in sequential image classification. The model trains on realistic elephant photos (Task 1) and subsequently on bird \textit{drawings} (Task 2), which typically causes it to forget the initial task. When Task 1 recurs as elephant \textit{drawings}, performance drops, illustrating the need to leverage task-invariant features to preserve knowledge during shifts.}
    \label{fig:introduction}
\end{figure}

Continual learning, also known as lifelong learning, plays a crucial role in enhancing the effectiveness of online prediction models. This concept facilitates the retention of previously learned information while integrating new insights, thereby maintaining a comprehensive understanding of the data distribution landscape. By leveraging task-specific and task-invariant features, continual learning helps to mitigate catastrophic forgetting. Fig.~\ref{fig:introduction} illustrates a stream of animal image classification tasks. Initially, the model classifies elephant photographs, followed by a task classifying bird drawings. Typically, an online model forgets how to classify elephants when switching to birds. In the performance depicted for the occurrences of \textit{Task 1}, we observe a decline after learning \textit{Task 2} when the model is compelled to deal again with the first task. However, by identifying task-invariant features, such as distinguishing between photographs and drawings, we have the potential to preserve valuable information for the subsequent reappearance of the initial task, for instance when involving elephants depicted in drawings. By understanding the distinctions between photographs and drawings and incorporating task-specific insights about elephants, we improve the model's ability to classify representations of elephants in drawings accurately. These features can be incorporated as prompts. A prompt is an additional piece of information provided to the model to guide its learning or inference process. By using prompts that highlight task-specific and task-invariant features, we can maintain the contextual knowledge necessary for effectively addressing evolving tasks, ensuring continued success in classification, particularly when faced with shifts in a task.

In this paper, we propose a framework called Continual Next Activity Prediction with Prompts (CNAPwP) with components for handling temporal dependencies, capturing long-term patterns, and incorporating memory and context into the learning process.  The key contributions of this work include: 

\begin{itemize}
    \item Introducing an innovative framework that combines general and expert prompts for dynamic adaptation and learning from streaming data.
    \item Developing the ``task-specific forgetting" evaluation metric to assess the retention and the degradation of task-specific knowledge.
    \item Creating new datasets that are specifically tailored for evaluating online predictive process monitoring.
    \item Conducting a comprehensive experimental evaluation that demonstrates that CNAPwP significantly outperforms existing methods in dynamic environments.
\end{itemize}

This paper extends, but fundamentally differs from our paper in \cite{verbeek2024handling}. Among others, this paper extends \cite{verbeek2024handling} by: (i) updating the related work, (ii) enriching the experimental evaluation with an extended ablation study, (iii) introducing a new real-world dataset (\texttt{BPIC2017}) and (iv) extensively evaluating the introduced ``task-specific forgetting" metric. 

The remainder of this paper is structured as follows: Section \ref{section:related} provides an overview of the state-of-the-art in continual learning and PPM. Section \ref{section:problem} introduces the foundational concepts, including concept drift and catastrophic forgetting, and formally defines the problem of task incremental learning for next activity prediction. Section \ref{section:main} details the proposed CNAPwP framework, explaining the architecture and its components for prompt-based adaptation. Section \ref{section:experimental_setup} outlines the experimental setup, encompassing datasets, baselines, and specific evaluation metrics used. Section \ref{sec:evaluation_results} presents a comprehensive analysis of the comparative results and the ablation study. Finally, Section \ref{section:conclusion} concludes the paper with a summary of the contributions and an outlook on future research.

%% file: related.tex
\vspace{-1mm}
\section{Related Work}
\label{section:related}
\vspace{-1mm}
At its core, continual learning seeks to equip machine learning models with the ability to continuously learn and adapt over time, similar to how humans acquire and refine knowledge throughout their lives. Within this framework, researchers have explored various methodologies, each offering unique perspectives and strategies. These approaches include memory-based, architecture-based, regularization based, and prompt-based methods.

\textbf{Memory-based approaches}, inspired by human cognition and memory systems, focus on storing and retrieving past experiences or knowledge. These methods can be broadly classified into two categories. The first involves retaining actual past experiences, as demonstrated by techniques such as Experience Replay \cite{rolnick2019experience}, iCaRL \cite{rebuffi2017icarl}, DynaTrainCDD \cite{kosciuszekonline}, Maximally Interfered Retrieval \cite{rahaf2019online}, and Gradient Episodic Memory \cite{lopez2017gradient}. The second category, on the other hand, involves generating past experiences during training, with Generative Replay \cite{lesort2019generative} serving as a key example.

\textbf{Regularization-based approaches} aim to prevent catastrophic forgetting by constraining weight updates during training. This constraint is typically achieved by evaluating the importance of each parameter for previously learned tasks. For example, Elastic Weight Consolidation \cite{kirkpatrick2017overcoming} determines parameter significance, while Synaptic Intelligence \cite{zenke2017continual} imposes penalties during the training of new tasks to preserve essential knowledge. Alternatively, the importance of parameters can be assessed based on their impact on output sensitivity, with selective penalties applied to critical parameters to mitigate forgetting, as seen in Learning without Forgetting \cite{li2017learning}.

\textbf{Architecture-based approaches} prioritize adjusting the neural network's structure to integrate new data while preserving existing knowledge. One approach involves dynamic architectures, which expand the network by adding more neurons or layers for each task. This allows the model to continuously grow and adapt without forgetting previous knowledge, as exemplified by methods such as Progressive Neural Networks \cite{rusu2016progressive} and Dynamic Neural Networks \cite{hou2018lifelong}.

\textbf{Prompt-based approaches}, a more recent addition to the continuum, introduce a novel perspective on continual learning challenges. These methods entail attaching static or adaptable “instructions”, also referred to as prompts, to direct the model's behavior. These prompts can take various forms, such as specific input patterns, embeddings, or task-specific tokens, and they help the model recall and apply knowledge from earlier tasks. Examples of prompt-based approaches include Learning to Prompt \cite{L2P} and DualPrompt \cite{wang2022dualprompt}.

We will delve deeper into DualPrompt \cite{wang2022dualprompt}, since our approach adopts a modified version of it. DualPrompt is a continual image classification method, inspired by prompting techniques commonly used in natural language processing to guide models toward desired results \cite{wang2022dualprompt}. This approach is particularly useful for sequence-like data, as it helps the model understand the underlying structure and the relationships within the sequences. It comprises two key components: G-Prompt and E-Prompt, which learn task-invariant and task-specific knowledge, respectively. The architecture of DualPrompt involves multiple multi-head self-attention (MHSA) layers, with G-Prompt and E-Prompt integrated into these layers. The G-Prompt is a shared parameter for all tasks, applied during both training and testing to capture general knowledge. The E-Prompt on the other hand contains a set of task-specific parameters, each linked to a learnable key representing the task's unique features. These prompts are strategically attached to different layers of the model, with G-Prompt typically at shallower layers and E-Prompt at deeper layers. This setup leverages the varying levels of feature abstraction in the layers. Two prompting functions can be used to combine the prompts with embedding features. In prompt tuning, prompts are attached to the input tokens of each layer, increasing the output length. In prefix tuning, the prompt is split and attached to specific tokens, maintaining the same output length as the input. During inference, an input is transformed to identify the closest task key and the associated E-Prompt using cosine similarity. The shared G-Prompt and matched E-Prompt are then attached to multiple MHSA layers to produce the model's output. During training, the E-Prompt and G-Prompt are jointly trained with the classifier based on the task identity.

TFCLPM \cite{TFCLPM} is the first continual learning method specifically designed for PPM. It utilizes a memory-based approach to store hard samples and a dynamic loss to predict next activities while mitigating catastrophic forgetting.

Outside continual learning, innovative methodologies have emerged in offline next-event prediction. The state of the art in offline next activity prediction is evolving through advancements in deep learning architectures, attention mechanisms, hybrid modeling approaches, and the integration of Generative Adversarial Networks (GANs), enhancing sequence generation and capturing complex temporal dependencies \cite{taymouri2020predictive}. Studies explore incremental techniques to update predictive models with new process execution data. Pauwels and Calders \cite{pauwels2021incremental} compare various update strategies, including re-training with and without hyperoptimization and incremental updates, demonstrating the effectiveness of incremental updates in maintaining model quality while offering real-time adaptability.

To efficiently store traces, prefix trees \cite{guzman2022log} offer a robust solution. These trees, represented as simple graphs, organize event sequences efficiently, with each node representing an event and edges connecting sequential events. As events occur, nodes are either created or their frequencies updated, capturing unique event sequences for individual cases.

Table \ref{tab:summary} summarizes the related work discussed in this paper from the prediction setup perspective as well as from the model architecture perspective.

\begin{table}[ht]
  \centering
  \rotatebox{270}{%
    \begin{minipage}{\textheight}
      \centering
      \caption{A summary table is provided, detailing all methods discussed in the preceding sections. Each method is classified according to its method type, which may include an image classification method, a general method focusing solely on the model updating, or a next activity prediction method (NA Pred). For each method several characteristics are highlighted, including whether the task-ID is unknown during testing, whether the task-ID is unknown during training, whether the method is a continual learning algorithm (CL), and the type of technique utilized (Memory-based, Regul.=Regularization-based, Arch.=Architecture-based, or Prompt-based).}
      \label{tab:summary}
      \begin{tabular}{llcccccccc}
        \toprule
        & \multirow{2}{*}{Methods} & \multicolumn{3}{c}{Settings} & \multicolumn{5}{c}{Techniques} \\
        \cmidrule(lr){3-5} \cmidrule(lr){6-10}
        & & \makecell{Task-Free\\ (Inference)} & \makecell{Task-Free\\ (Training)} & \makecell{Continual\\ Learning} & \multicolumn{2}{c}{Memory-Based} & Regularization & Architecture & Prompt-Based \\
        \cmidrule(lr){6-7}
        & & & & & Raw & Pseudo & & & \\
        \midrule
        \multirow{12}{*}{\rotatebox[origin=c]{90}{Image Classification}}
          & Generative Replay \cite{shin2017continual}                                          & \checkmark & \checkmark & \checkmark &            & \checkmark &            &            &            \\
          & Maximally Interfered Retrieval \cite{rahaf2019online}                              & \checkmark & \checkmark & \checkmark & \checkmark &            &            &            &            \\
          & Experience Replay \cite{rolnick2019experience}                                     & \checkmark & \checkmark & \checkmark & \checkmark &            &            &            &            \\
          & iCaRL \cite{rebuffi2017icarl}                                                      & \checkmark & \checkmark & \checkmark & \checkmark &            &            &            &            \\
          & Progressive Neural Networks \cite{rusu2016progressive}                             & \checkmark & \checkmark & \checkmark &            &            &            & \checkmark &            \\
          & Dynamic Neural Networks \cite{rusu2016progressive}, \cite{hou2018lifelong}         & \checkmark & \checkmark & \checkmark &            &            &            & \checkmark &            \\
          & Learning without Forgetting \cite{li2017learning}                                  & \checkmark & \checkmark & \checkmark &            &            & \checkmark &            &            \\
          & Elastic Weight Consolidation \cite{kirkpatrick2017overcoming}                      & \checkmark & \checkmark & \checkmark &            &            & \checkmark &            &            \\
          & Gradient Episodic Memory \cite{lopez2017gradient}                                  & \checkmark & \checkmark & \checkmark &            &            & \checkmark &            &            \\
          & Synaptic Intelligence \cite{zenke2017continual}                                    & \checkmark & \checkmark & \checkmark &            &            & \checkmark &            &            \\
          & Learning to Prompt \cite{L2P}                                                      & \checkmark & \checkmark & \checkmark &            &            &            &            & \checkmark \\
          & DualPrompt \cite{wang2022dualprompt}                                               & \checkmark & \checkmark & \checkmark &            &            &            &            & \checkmark \\
        \midrule
        \multirow{5}{*}{\rotatebox[origin=c]{90}{\makecell{Updating\\ Strategies}}}
          & Incremental Update \cite{pauwels2021incremental}                                   & \checkmark & \checkmark & \checkmark & \checkmark &            &            &            &            \\
          & Full Re-train \cite{pauwels2021incremental}                                        & \checkmark & \checkmark & \checkmark & \checkmark &            &            &            &            \\
          & \makecell[l]{Re-train with\\No Hyperoptimization} \cite{pauwels2021incremental}   & \checkmark & \checkmark & \checkmark & \checkmark &            &            &            &            \\
          & Do Nothing \cite{pauwels2021incremental}                                           & \checkmark & \checkmark & $\times$   &            &            &            &            &            \\
          & ADWIN \cite{bifet2007learning}                                                     & \checkmark & \checkmark & \checkmark & \checkmark &            &            &            &            \\
        \midrule
        \multirow{3}{*}{\rotatebox[origin=c]{90}{\makecell{Next\\ Activity}}}
          & DynaTrainCDD \cite{kosciuszekonline}                                               & \checkmark & \checkmark & \checkmark & \checkmark &            &            &            &            \\
          & TFCLPM \cite{TFCLPM}                                                               & \checkmark & \checkmark & \checkmark & \checkmark &            & \checkmark &            &            \\
          & \makecell[l]{GAN} \cite{taymouri2020predictive}   & \checkmark & $\times$   & $\times$   &            & \checkmark &            &            &            \\
        \bottomrule
      \end{tabular}
    \end{minipage}%
  }
\end{table}

%% file: problem.tex
\section{Preliminaries and Problem Formulation}
\label{section:problem}
In this section we delve into formulating the problem of utilizing continual learning in process prediction. We aim to develop an algorithm that continuously processes an event stream $S =\{ e_1, e_2, ... \}$ as events are generated, where an event $e = (c, a, t, v_1, ..., v_{\mathcal{A}})$ is a tuple of case identifier $c$, activity label $a$, timestamp $t$, and the values of other event attributes $v_1, ..., v_{\mathcal{A}}$. A case refers to a single instance of the process being analyzed or executed, encompassing all events, their attributes, and contextual information associated with that instance. A trace $\sigma^{(i)} = \langle e_{1i}, ..., e_{ni} \rangle$ denotes a finite sequence of events that all belong to the same case $i$ (i.e. they share the same case identifier $c$), ordered by their timestamps. In our notation, a single subscript (e.g. $e_t$ in stream $S$) denotes the event's sequential position in the global stream. A double subscript (e.g. $e_{ji}$ in trace $\sigma^{(i)}$) denotes the $j$-th event occurring within the specific case $i$. Without loss of generality and since we are dealing merely with event logs, the terms trace and case are used interchangeably in the remainder of the paper and refer to the sequence of activities within the instance. A stream $S$ contains multiple traces. 

Given $\sigma^{(i)}$, a prefix represents the sequence of activities executed up to a certain point in a trace's lifecycle. The prefix of length $k$ is defined by $\sigma_{\leq k}^{(i)} = \langle e_{1i}, ..., e_{ki} \rangle$. For each event $e$ that occurs in stream $S$, we want to predict the next activity name of event $e_{(k+1)i}$.

\begin{definition}[Next Activity Prediction]\label{next_act}
    Let there be a sample of
    prefixes of sequences $\mathcal{P} = \{\sigma_{\leq k}^{(i)}\}_{i = 1}^{i = m}$ where $2 \leq k < |\sigma^{(i)}|$ is the prefix length and $m$ is the sample size. Given a prefix of an events sequence $\sigma_{\leq k}^{(i)}$, the \textit{next activity prediction} is $\hat{a}_{(k+1)i}$ of activity $a_{(k+1)i}$.
\end{definition}

To continuously predict the next activity in a stream, we must engage in online next activity prediction.

\begin{definition}[Online Next Activity Prediction]\label{online}
Assume we aim to perform ongoing predictions of the next activity on a stream of events. Upon the arrival of each new event $e_{ki}$, we continuously utilize the prefix $\sigma_{\leq k}^{(i)}$ to predict the subsequent activity $a_{(k+1)i}$.
\end{definition}

A stream $S$ encompasses multiple learning tasks, each having several traces.
\begin{definition}[Learning Task]\label{learn_task}
A model will learn a \textit{learning task} $\mathcal{T}_n$ with a corresponding dataset $\mathcal{D}_n$ for $n \in [1, ..., \mathcal{N}]$ where $\mathcal{N}$ is the number of learning tasks. Let $\mathcal{T}_p$ and $\mathcal{T}_r$ represent two distinct learning tasks, where $p \neq r$. This indicates that $\mathcal{T}_p$ and $\mathcal{T}_r$ belong to different processes or data , implying there is no relationship between these learning tasks. 
\end{definition}

In the context of a stream of events involving multiple learning tasks, the distribution evolves over time, necessitating continuous updates to the prediction model. This demands a special form of incremental learning, known as \textit{task incremental learning.}

\begin{definition}[Task Incremental Learning]\label{task_inc}
In \textit{task incremental learning}, two tasks $\mathcal{T}_n, \mathcal{T}_m$ where $n, m \in [1, ..., \mathcal{N}]$ have no correspondence to each other if $n \neq m$. Each task has unique objectives and is associated with a separate process, necessitating the model to acquire new patterns, features, or behaviors.
\end{definition}

The emergence of new tasks results from concept drifts.
\begin{definition}[Concept Drift \cite{gama2014survey}]\label{concept}
A concept drift between timestamps \( t_0 \) and \( t_1 \) is defined as $\exists X : p_{t_0}(X, y) \neq p_{t_1}(X, y)$ where \( p_{t_0} \) denotes the joint distribution at time \( t_0 \) between the set of input variables \( X \) and the target variable \( y \). 
\end{definition}
A task repeating multiple times is called a recurrent concept drift.

\begin{definition}[Recurrent Concept Drifts]\label{recurrent_drift}
 A recurrent concept drift implies repeated changes in tasks, with the possibility of cycles that bring the system back to the original task.   
\end{definition}

When updating the model after a concept drift, the goal is to utilize as much data as possible. If a task reappears, it is advantageous to have stored data about this task, ensuring that it is not forgotten. However, this poses challenges such as storage limitations or privacy concerns. This often prevents keeping all past data, but relying only on recent data introduces bias and causes catastrophic forgetting \cite{chrysakis2020online}.

\begin{definition}[Catastrophic Forgetting \cite{kirkpatrick2017overcoming}]\label{cata}
    Let there be a model at any point in time that has learned a sequence of $\mathcal{T}$ learning tasks. When faced with the $(\mathcal{T} + 1)$th task, the model tends to forget how to predict the next activities of the $\mathcal{T}$ previously learned tasks.
\end{definition}

This phenomenon poses a significant challenge in dynamic environments. To address the issue of catastrophic forgetting, continual learning enables models to retain and incorporate knowledge from previous tasks while learning new ones. 

\begin{definition}[Continual Learning \cite{kosciuszekonline}]\label{continual_learn}
    Let there be a model at any point in time that has learned a sequence of $\mathcal{T}$ learning tasks. When faced with the $(\mathcal{T} + 1)$th task, the model can leverage the past knowledge in the knowledge base to help learn this task. The objective is to optimize the performance of the new task while minimizing the decrease in performance on the previous tasks.
\end{definition}
By compiling the definitions, we arrive at the following definition for a continual learning algorithm used to predict the next activity.

\begin{definition}[Continual Learning for Next Activity Prediction]\label{cont_next}
    Consider a model that has learned to predict the next activities for an event stream $S$. Throughout $S$, concept drifts may occur which refer to the alteration in the learning task. The event stream contains a sequence of $\mathcal{T}$ learning tasks. A learning task for next activity prediction is defined by a sample of prefixes of sequences $\mathcal{P}$. Given a prefix of an events sequence $\sigma_{\leq k}^{(i)}$, the model has to predict the activity $\hat{a}_{(k+1)i}$. When presented with the $(\mathcal{T} + 1)$th task, the model must accurately predict the activities for this new task without suffering from catastrophic forgetting.
\end{definition}

The solution to the challenges of continual learning in process prediction models must meet several key requirements. Firstly, it should efficiently utilize an input event stream $S$ consisting of multiple learning tasks $\mathcal{T}$, as defined in Definition \ref{learn_task}, and predict next activities accurately in near-real-time, as described in Definition \ref{online}. Secondly, the solution should accommodate task incremental learning, mentioned in Definition \ref{task_inc}, allowing for seamless integration in the emergence of new tasks. Additionally, the solution must exhibit the capability to perform next activity predictions while remaining robust to concept drifts, mentioned in Definition \ref{concept}. Moreover, the solution should aim to reduce catastrophic forgetting, as defined in Definition \ref{cata}, particularly when tasks recur in the stream, as defined in Definition \ref{recurrent_drift}. In essence, the algorithm should incorporate continual learning for predicting next activities, as defined in Definition \ref{cont_next}.

%% file: main.tex
\section{Method}\label{section:main}
\begin{figure*}[h]
\centering
\resizebox{\columnwidth}{!}{
\includegraphics{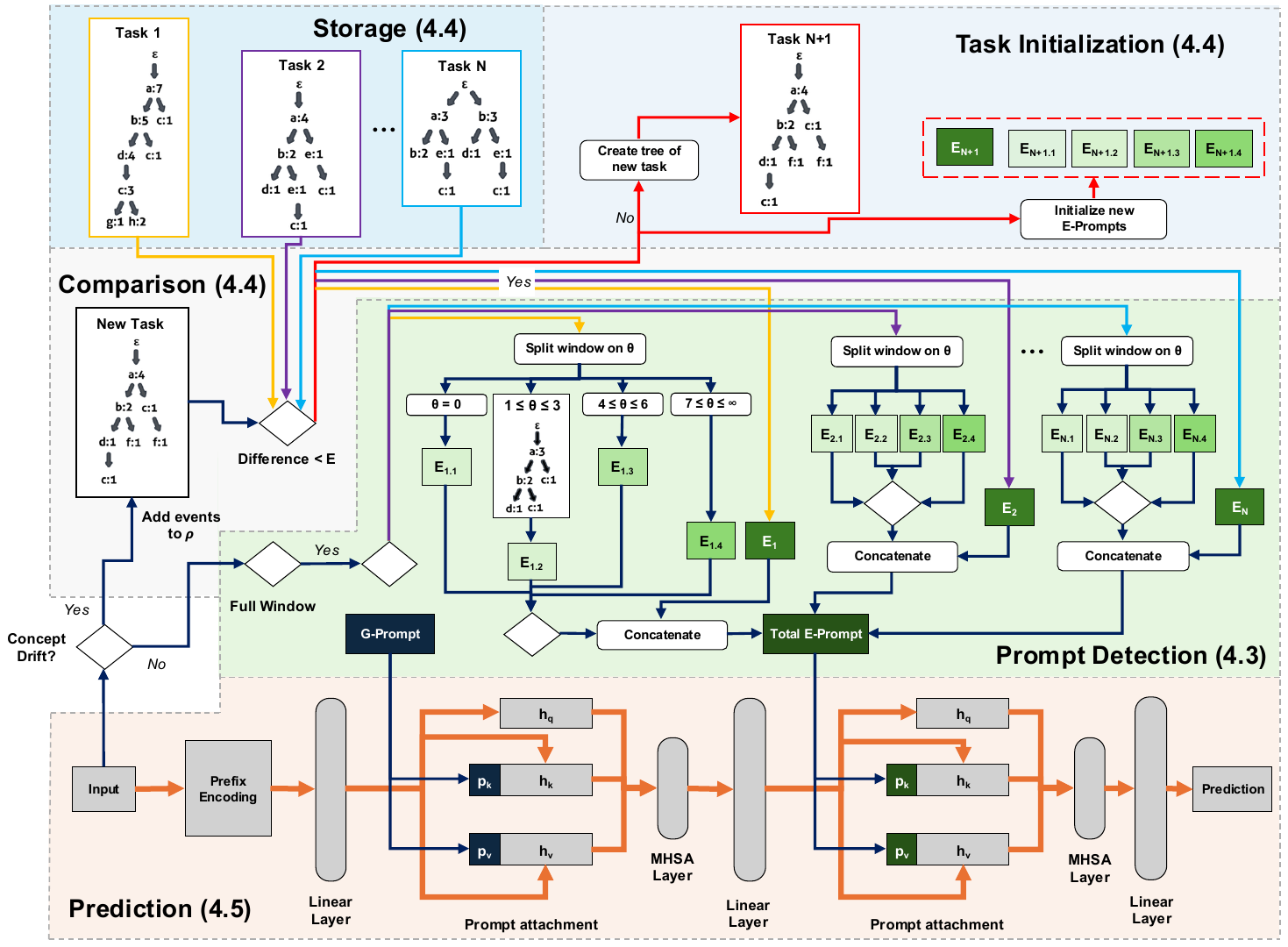}}
\caption{Framework for Continual Next Activity Prediction with Prompts (CNAPwP). When concept drift is \textit{externally} detected, a new task-specific E-Prompt is initialized. When no drifts are detected, a prompt detection mechanism dynamically selects and attaches the appropriate general G-Prompt and E-Prompt to the MHSA layers for accurate prediction.}
\label{fig:arch}
\end{figure*}

In this section, we provide an overview of the CNAPwP architecture by detailing its components. The implementation is available on GitHub\footnote{https://github.com/SvStraten/CNAPwP}. Fig.~\ref{fig:arch} shows the framework architecture for training and testing the prediction model. First, raw event streams undergo preprocessing (Section~\ref{subsec:preprocessing}) to generate and encode prefixes. These processed events  are managed by the window handling component (Section~\ref{subsec:window_handling}), which maintains a dynamic sliding window of the most recent data. This window feeds into the window updating and prompt detection (Section~\ref{subsec:window_updating}) mechanism, which continuously updates the model parameters and selects the appropriate general prompt (G-Prompt) for shared knowledge and expert prompt (E-Prompt) for task-specific knowledge. Simultaneously, the concept drift detection (Section~\ref{subsec:concept_drift_detection}) component monitors the stream for distributional shifts, triggering the storage of new task definitions and the initialization of new prompts when a new task is identified. Finally, the prediction model (Section~\ref{subsec:prediction_model}) integrates the input prefix with the selected prompts via a multi-head self-attention (MHSA) architecture to output the probability distribution for the next activity.

\subsection{Preprocessing}\label{subsec:preprocessing}
Preprocessing involves initial data reduction, where unnecessary columns are removed from the event log to reduce noise, focusing on essential attributes such as case, event, resource, and completion time. If events are missing key attributes, the corresponding cases are excluded from the dataset. Attributes representing the resource should be retained if present. Next, data is ordered chronologically to represent a real-time data stream, and afterward, the completion time attribute is deleted.  Next, for each new event, a prefix is generated. The approach checks whether the case has appeared within the window of events. If the case is found and there are sufficient events to match the prefix size, the prefix is formed. To accommodate the variable nature of trace lengths, the framework supports different prefix sizes $k$. While all prefixes are padded to a fixed maximum sequence length to facilitate tensor operations and batching within the neural network, the effective length $k$ of each prefix is preserved. This effective length is subsequently used to assign a specific ``bucket" (as detailed in Section~\ref{subsec:window_updating}), ensuring that the model applies the appropriate length-dependent prompts during training and inference. If the case is new and not previously encountered, a prefix composed entirely of ``None'' values is created. Once the prefix is defined, the next activity is assigned as the new event. Events outside the defined window are disregarded in this process.  After generating the prefixes, the next step involves encoding both the prefix and the next activity. The prefix is one-hot encoded, generating sparse input vectors that are effective for handling high-dimensional categorical data, whereas the next activity is ordinally encoded, offering a compact and efficient representation.

\subsection{Window Handling}\label{subsec:window_handling}
The process starts by importing an event log and extracting key attributes necessary for predicting the next activity. These attributes are converted into a continuous event flow, transmitting events until the entire log is analyzed. Events are sequenced by activity completion order, allowing new cases to begin while others are still ongoing. As each event arrives, it is added to a dynamic window that accumulates events until it reaches its window size, defined by $\gamma$. The optimal value of this parameter is established based on the experiment outlined in Section \ref{parameters}. Once full, the window operates on a ``first-in, first-out" basis, maintaining a consistent size to manage the sequence of events effectively.

\subsection{Window Updating and Prompt Detection}\label{subsec:window_updating}
When $\gamma$ events accumulate, an update is initiated, as depicted in the green segment of Fig.~\ref{fig:arch}. This ensures the model remains current with the evolving event stream, maintaining a dynamic framework ready to process new events seamlessly. After accumulating $\gamma$ events, the data is partitioned into buckets based on prefix lengths $\theta$. Each bucket represents a specific range of prefix lengths, ensuring a uniform distribution. Fig.~\ref{fig:arch} illustrates that there are four buckets, with the first containing solely empty prefixes and the second containing prefixes ranging in length from $1$ to $3$, for example. During initialization, bucket ranges are selected based on optimal values. Data of a specific prefix length is added to buckets until a significant discrepancy occurs, ensuring each bucket has a comparable number of values.  With the window data segmented into buckets, the next phase updates all method parameters, including the G-Prompt, the bucket-specific E-Prompts, the task-specific E-Prompt, and the model weights. Data is batched by bucket, with each batch containing data from a single bucket. Data from each bucket is used solely for updating the parameters within its corresponding bucket-specific E-Prompt. The task-specific E-Prompt, G-Prompt, and weights receive updates across all batches. The cross-entropy loss is used for updating the weights and parameters of the G-Prompt and E-Prompt:

\begin{equation}
\text{CrossEntropyLoss} = -\sum_{i=1}^{N} y_i \log(\hat{y}_i)
\end{equation}

where $N$ is the number of classes, $y_i$ and $\hat{y}_i$ are the true and predicted probabilities for class $i$, respectively. Both the G-Prompt and E-Prompt are concatenated with the model to facilitate these targeted updates. An ablation study experiment, detailed in Section \ref{ablation}, is conducted to validate the use of both the G-Prompt and E-Prompt.

\subsection{Concept Drift Detection}\label{subsec:concept_drift_detection}
Routine weight and parameter updates continue until a concept drift is detected. CNAPwP assumes these shifts are anticipated, triggering a task recognition mechanism upon encountering a drift index. After a concept drift, a specified number of events, denoted as the buffer size $\rho$, is stored in a prefix tree as depicted in the grey segment of Fig.~\ref{fig:arch}. The optimal value of $\rho$ is established through an experiment outlined in Section \ref{parameters}. The prefix tree of the new task is compared with the prefix trees of stored tasks, as shown in the blue section of Fig.~\ref{fig:arch}. The comparison focuses on identifying overlaps and unique sequences between them. A high degree of overlap and fewer unique sequences in the new task’s prefix tree indicate greater similarity between the trees. A dissimilarity ratio is then calculated to measure how distinct the tasks are, with the lowest ratio indicating the stored task most similar to the new task. For each dataset, a threshold $\mathcal{E}$ is established according to the dissimilarity between its constituent tasks, . If the similarity between the new task and any of the stored tasks is less than $\mathcal{E}$, the model will utilize the E-Prompts associated with that task. The optimal value for $\mathcal{E}$ for each dataset is determined in an experiment outlined in Section \ref{parameters}.

If none of the stored tasks are similar to the new task, our model does not overfit to only previously seen tasks and, as such, initializes parameters for the new task, as depicted in the pink segment of Figure \ref{fig:arch}. This involves setting up the E-Prompt parameters for the new task and each bucket, which are initialized randomly. A new prefix tree is also created to characterize the task and is stored together with the tasks already in storage. As events stream in, they are added to the prefix tree until it reaches 500 events, enough to capture the task's essence for future comparisons.

\subsection{Prediction Model}\label{subsec:prediction_model}
A multi-head self-attention (MHSA) mechanism is utilized for next activity predictions. The MHSA mechanism enhances next activity prediction by capturing complex dependencies and relationships between events, allowing the model to weigh the significance of past activities effectively. This approach enables the model to handle variable-length sequences efficiently while extracting diverse features, leading to more accurate predictions based on historical patterns. The prediction model of the approach is depicted in the orange segment of Figure \ref{fig:arch}. 

The prefix of each input event from the event stream is used as the model input. During preprocessing, each event's prefix is converted into a one-hot encoded array with dimensions $[k, num\_events]$, where $k$ is the prefix length and $num\_events$ is the number of distinct events. Therefore, the encoded prefix of the event is initially determined. Subsequently, this encoded prefix undergoes a series of layers, ultimately resulting in a prediction of the next activity. The model consists of duplicated linear layers and MHSA layers. Linear layers take inputs of size $input\_size$ and produce outputs of $3 \times input\_size$, allowing prompt attachment via prefix tuning. For further details on prefix tuning, please refer to \cite{wang2022dualprompt}. These outputs are then processed by a MHSA layer with softmax activation and dropout regularization. Outputs from the MHSA layers feed into a dense layer with matching input and output sizes, equivalent to the number of events times the prefix length. Finally, the output passes through a softmax layer, converting it into a probability distribution over possible activities. The activity with the highest probability is predicted as the next activity.

During inference, the G-Prompt and E-Prompt are appended to the input of the MHSA layers. The G-Prompt remains consistent across all inputs, while the E-Prompt is determined by the task recognition mechanism. The task linked with the input event is identifiable due to the mechanism activated after a concept drift, as explained in Section \ref{subsec:concept_drift_detection}. The bucket-specific E-Prompt is chosen based on the prefix size $k$ of the input event. This bucket-specific E-Prompt is combined with the task-specific E-Prompt and applied to the input using a designated prompting function, such as prefix tuning. The justification of selecting the prompting function is based on the related experiment described in Section \ref{ablation}. 

%% file: experiments.tex
\section{Experimental Setup}
\label{section:experimental_setup}

\subsection{Evaluation Metrics}
Typically, studies rely on metrics such as accuracy and running time to assess performance. However, these metrics often fall short of capturing the longitudinal performance of our model, which can vary significantly over time. We will explore alternative evaluation metrics that offer a more nuanced understanding of our model's performance dynamics over extended periods.

\textbf{Accuracy at a given event index} is a performance metric used to assess the effectiveness of a predictive model at a specific position within the sequence of events. In mathematical terms, if we denote $w$ as the size of the average accuracy window, and \( \hat{y}_j \) as the predicted activity label of an event at index \( j \) based on preceding events, with \( y_j \) representing the actual activity at index \( j \), the accuracy at index \( i \) can be expressed as:
  $$  accuracy_i = \frac{1}{w} \sum_{j=i-w}^{i} \text{1}\{ \hat{y}_j = y_j \} $$
The formula computes the ratio of correct predictions from \( j \) up to event index \( i \), averaging the predictions in the window.

\textbf{Average accuracy} is calculated by averaging across all events to determine if each activity is predicted correctly. To calculate the average accuracy, we have the following formula:
 $$accuracy = \frac{1}{\mathcal{N}} \sum_{j=0}^{\mathcal{N}} \text{1}\{ \hat{y}_j = y_j \} $$
Where \( \mathcal{N} \) is the total number of events in the stream. In both cases, \( \text{1}\{\cdot\} \) denotes the indicator function, yielding 1 if its argument holds and 0 otherwise.

\textbf{Task-specific forgetting} is the difference in the accuracy of predictions when the task is encountered for the first time compared to the accuracy during subsequent encounters. For each task and each appearance of a task, we compute the following:
  $$  task\_accuracy(n, a) = \mathcal{R}_{t_{n},t_{n1}} - \mathcal{R}_{t_{n},t_{na}} $$
Where \( n \) is the task, \( \mathcal{R}_{t_n, t_{ni}} \) is the accuracy of task \( t_n \) after encountering it for the \( i^{th} \) time, and \( a > 1 \) is the appearance of the subsequent task, except for the first task. 

The \textbf{ running time} is computed by taking the difference between the end time and the start time of the full evaluation stream, incorporating both training and inference.

\subsection{Datasets}
We provide an overview of the datasets utilized in our study, comprising a diverse selection of both real-world and synthetic datasets. 

To generate sudden concept drift datasets, the following scheme was employed: for each concept, \( \mathcal{N} \) examples are sampled. The samples of each concept are concatenated to form a new event log, alternating between the concepts. Consequently, a sudden concept drift is introduced at every \( \mathcal{N} \) events.

To replicate the occurrence of a drift in a log, the authors of \cite{maaradji2015fast} created a reference set of 72 event logs by modifying various parameters. They systematically modified a base model by applying one of twelve simple change patterns. 
Using the Business Process Drift data, three datasets were generated, each with different lengths and characteristics. The first dataset, \texttt{RandomTasks}, consists of seven distinct tasks arranged in batches without any specific order. Each batch contains all tasks, and the entire dataset is made up of four batches, totaling \( 80,406 \) events. The second dataset, \texttt{ImbalancedTasks}, includes four tasks that occur with varying frequencies, with each task appearing between one and four times, totaling \( 30,754 \) events. The third dataset, \texttt{RecurrentTasks}, features four distinct tasks that recur within the same loop multiple times, totaling \( 100,739 \) events.

In terms of real-world datasets, we utilize a dataset named \texttt{Recurrent BPIC2015}, which is a recurrent dataset generated from actual data collected across five different municipalities. Each task within BPIC2015 corresponds to data from a specific municipality. This \texttt{Recurrent BPIC2015} dataset comprises \( 32,016 \) events. Additionally, we include the \texttt{BPIC2017} dataset. This dataset originates from loan application processes within a Dutch financial institution. It contains events representing the full lifecycle of loan applications, including application submission, document verification, and final approval or rejection. The dataset exhibits high variability in case lengths and contains multiple parallel and conditional paths, making it suitable for evaluating complex, real-world dynamics.

\subsection{Baselines}
The model's performance is compared against five competitors, including two baseline methods: Landmark and Incremental Update (w = Last Drift). In the \textbf{Landmark} approach \cite{pauwels2021incremental}, the model retrains from scratch with each new data window. In the \textbf{Incremental Updating (w = Last Drift)} approach \cite{pauwels2021incremental}, the model updates based on historical data up to the last observed concept drift.

In addition to these baselines, three state-of-the-art next activity prediction methods are used as competitors. \textbf{DynaTrainCDD} \cite{kosciuszekonline} which distinguishes itself through its advanced concept drift detection algorithm called PrefixTreeCDD \cite{guzman2022log}. It continually monitors process data for deviations and utilizes prefix trees to represent and analyze process sequences efficiently. These detected drifts dynamically dictate the frequency of updates and the selection of datasets for retraining. \textbf{TFCLPM} \cite{TFCLPM} dynamically updates its model based on a retraining dataset that combines recent events with hard samples. It is processed through a Single Dense Layer (SDL) \cite{pauwels2021incremental} neural network using a dynamic loss function that incorporates  Memory Aware Synapses (MAS) \cite{aljundi2018memory} to mitigate significant parameter changes. Finally, an approach that uses \textbf{GANs} \cite{taymouri2020predictive} is constructed upon the principle of establishing a competitive game between two entities, each represented by a Recurrent Neural Network (RNN). Throughout the training process, one player gradually learns to generate event sequences that closely resemble those observed in the training data, while the other player evaluates the realism of this prediction.

To ensure a fair comparison of the continual learning strategies only and not the utilized prediction models, both Landmark and IncrementalUpdate (w = LastDrift) are implemented using the same multi-head self-attention backbone architecture employed by CNAPwP. This isolates the impact of the update strategy from the model architecture. DynaTrainCDD \cite{kosciuszekonline} and TFCLPM \cite{TFCLPM} both use the SDL neural network, as proposed in their paper. The GAN-based approach \cite{taymouri2020predictive} relies on its specific adversarial architecture. Unlike other methods that utilize standard supervised loss functions (cross-entropy loss), the GAN-based approach relies on adversarial loss which is notoriously unstable during the initial training phases. Taymouri et al. \cite{taymouri2021deep} explicitly requires 500 training iterations to ensure the generator and discriminator reach a stable equilibrium. To adapt this offline requirement to our online setup and prevent immediate model collapse, the GAN model is initially trained on a warm-up set of 500 events before switching to the online setup.

\subsection{Implementation Details}\label{sec:implementation_details}
We implemented the framework using Python and the PyTorch library for neural network components. Data preprocessing was handled using Pandas and Numpy, while Seaborn and Matplotlib were used for visualizations. We conducted the experiments on an HP Envy x360 laptop equipped with an AMD Ryzen 5 2500U processor (2.00 GHz), 8.00 GB of RAM, and running Windows 11. The implementation is open source and available on GitHub\footnote{https://github.com/SvStraten/CNAPwP}.

To ensure a robust evaluation, we employ a temporal split of the event stream. The first 15\% of the data constitutes the validation set used for hyperparameter selection (as detailed in Section~\ref{parameters}), while the remaining 85\% is used for the evaluation. On this evaluation stream, we adopt a test-then-train protocol. This implies that each incoming event is first used to evaluate the model's prediction performance before it is used to train the model, ensuring that the model is always tested on unseen data. For the online training updates, we utilize a batch size of 25, a learning rate of 0.01 and perform 10 retraining epochs per update window.

\subsection{Parameter Selection}\label{parameters}
Three key hyperparameters are fine-tuned for CNAPwP, corresponding directly to the variables in Section~\ref{section:main}. The window size ($\gamma$) specifies the capacity of the sliding window used for event accumulation. The buffer size ($\rho$) represents the number of events stored in the prefix tree immediately following a concept drift detection. Finally, the threshold ($\mathcal{E}$) defines the dissimilarity limit used to determine whether new data corresponds to an existing task or requires the initialization of a new task. All parameter configurations were rigorously assessed by performing grid search on the validation set, using average accuracy and running time as our metrics.

\begin{table}[t]
    \centering
    \caption{The final parameter settings for all datasets.}
    \resizebox{0.8\columnwidth}{!}{
        \begin{tabular}{lccc}
            \toprule
            Dataset & \texttt{window\_size $\gamma$} & \texttt{buffer\_size $\rho$} & \texttt{threshold $\mathcal{E}$} \\
            \midrule
            \texttt{RandomTasks}       & 250 & 100 & 0.5 \\
            \texttt{ImbalancedTasks}   & 500 & 100 & 0.6 \\
            \texttt{RecurrentTasks}    & 250 & 150 & 0.6 \\
            \texttt{Recurrent BPIC2015}& 250 & 100 & 0.8 \\
            \texttt{BPIC2017}          & 500 & 100 & 0.5 \\
            \bottomrule
        \end{tabular}
    }
    \label{tab:params}
\end{table}

For grid search, we used $\gamma\in \{250, 500, 1000\}$, $\rho \in \{50, 100, 150\}$ and $\mathcal{E} \in \{0.2, 0.4, 0.5, 0.6, 0.8\}$. For \texttt{RandomTasks}, \texttt{RecurrentTasks}, \texttt{BPIC2017} and \texttt{Recurrent BPIC2015}, a $\gamma$ of 250 was selected as it provided the shortest running time while maintaining high accuracy. However, for the \texttt{ImbalancedTasks} dataset, a larger window size of 500 was necessary to maintain performance, ensuring sufficient data collection for infrequent tasks. $\rho$ and $\mathcal{E}$ control the sensitivity of task recognition. We observed that increasing $\rho$ often led to a drop in accuracy due to the extended period of uncertainty while filling the buffer. Therefore, we prioritized smaller buffer sizes (e.g. 100 or 150) to ensure rapid task identification.

\section{Evaluation Results}\label{sec:evaluation_results}
\subsection{Results}

\subsubsection{Average accuracy} 
Table~\ref{tab:accs} provides an overview of the average accuracy across all evaluation datasets and methods. The results indicate that different methods excel depending on dataset characteristics. TFCLPM \cite{TFCLPM} outperforms all other methods on the \texttt{ImbalancedTasks} and \texttt{BPIC2017} datasets, demonstrating its effectiveness in scenarios with fewer tasks or less frequent drifts. However, CNAPwP (ours) achieves the highest average accuracy on the \texttt{RandomTasks} and \texttt{RecurrentTasks} datasets, outperforming all other baseline methods. This suggests that our prompt-based approach is particularly effective in scenarios with structured recurrence or randomized batches of tasks.

\begin{table}[H]
    \centering
    \caption{Average accuracy and standard deviation (after $5$ runs over each evaluation
    set). \textbf{Bold} indicates the highest accuracy, while \textit{italic} and 
    \underline{underlined} second- and third-highest values, respectively.}
    \resizebox{\columnwidth}{!}{
        \begin{tabular}{lccccc}
            \toprule
            Method & \multicolumn{5}{c}{Dataset} \\
            \cmidrule(lr){2-6}
            & \texttt{RandTasks} & \texttt{ImbalTasks} & \texttt{Recur BPIC15} & \texttt{RecurTasks} & \texttt{BPIC17} \\
            \midrule
            CNAPwP (ours)          & \textbf{.813}{\tiny$\pm$.004} & \textit{.797}{\tiny$\pm$.007} & \textit{.662}{\tiny$\pm$.003} & \textbf{.789}{\tiny$\pm$.002} & \underline{.818}{\tiny$\pm$.002} \\
            DynaTrainCDD \cite{kosciuszekonline} & .747{\tiny$\pm$.005} & .758{\tiny$\pm$.006} & .543{\tiny$\pm$.007} & .731{\tiny$\pm$.004} & \textit{.820}{\tiny$\pm$.004} \\
            TFCLPM \cite{TFCLPM}   & \underline{.795}{\tiny$\pm$.001} & \textbf{.817}{\tiny$\pm$.004} & \underline{.655}{\tiny$\pm$.002} & \textit{.781}{\tiny$\pm$.001} & \textbf{.839}{\tiny$\pm$.000} \\
            GAN \cite{taymouri2020predictive} & .739{\tiny$\pm$.001} & .704{\tiny$\pm$.004} & .604{\tiny$\pm$.005} & .694{\tiny$\pm$.001} & .693{\tiny$\pm$.000} \\
            Landmark \cite{pauwels2021incremental} & \textit{.800}{\tiny$\pm$.007} & \underline{.796}{\tiny$\pm$.007} & \textbf{.670}{\tiny$\pm$.006} & \underline{.780}{\tiny$\pm$.002} & .798{\tiny$\pm$.004} \\
            w = LastDrift \cite{pauwels2021incremental} & .760{\tiny$\pm$.001} & .745{\tiny$\pm$.008} & .570{\tiny$\pm$.004} & .727{\tiny$\pm$.002} & .798{\tiny$\pm$.005} \\
            \bottomrule
        \end{tabular}
    }
    \label{tab:accs}
\end{table}

\subsubsection{ImbalancedTasks dataset} 
\begin{figure}[H]
    \centering
    \resizebox{\columnwidth}{!}{
\includegraphics{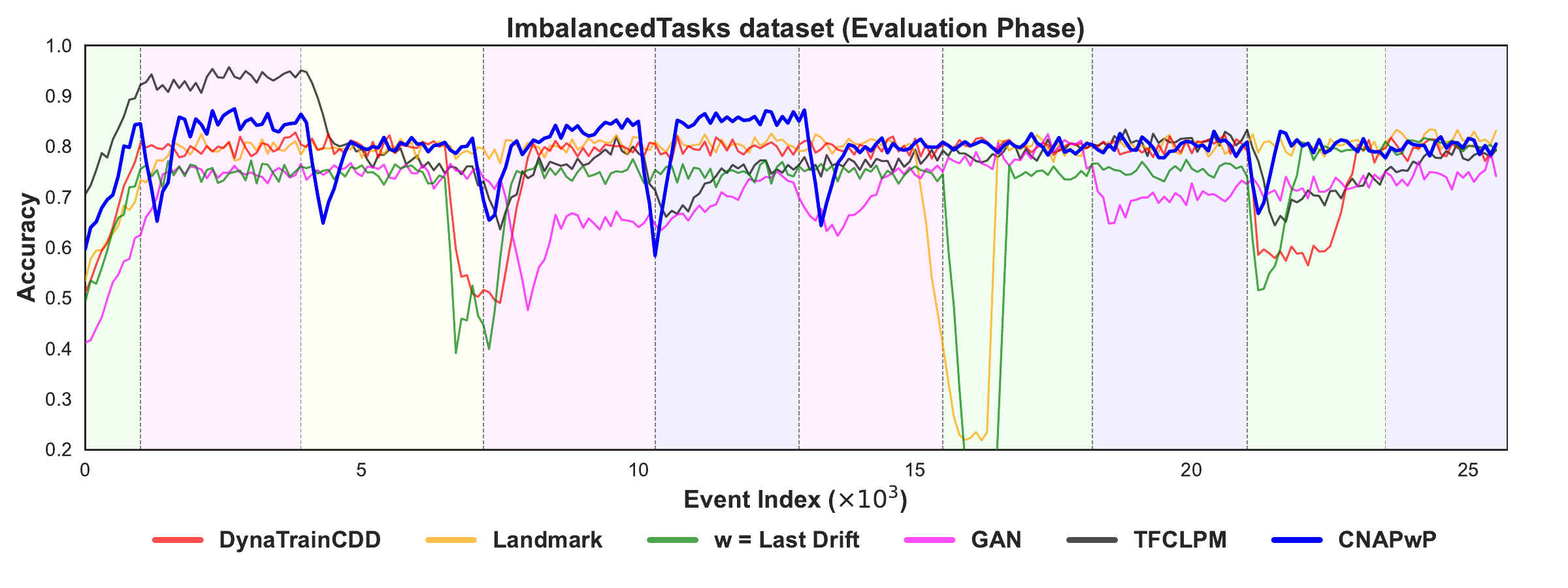}
        }
    \caption{The accuracy per event for all methods on the \texttt{ImbalancedTasks} dataset. The plot reveals that while baselines like TFCLPM initially perform well, they suffer from degradation in later stages of the stream. In contrast, CNAPwP (dark blue line) demonstrates robust adaptation, maintaining higher stability and positive performance deltas even as task frequencies vary significantly throughout the stream.}
    \label{fig:imbalanced_result}
\end{figure}

\begin{figure}[H]
    \centering
    \resizebox{\columnwidth}{!}{
\includegraphics{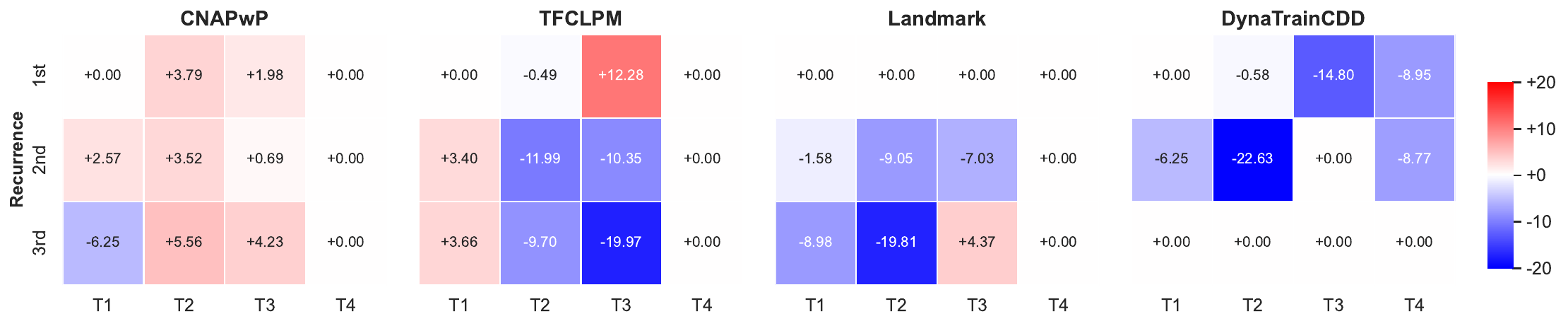}
        }
    \caption{Heatmaps showing the accuracy drop per task (task-specific forgetting) for the \texttt{ImbalancedTasks} dataset for the four best-performing methods. Warm colors indicate positive knowledge retentions (improvements), white indicates stability and cold colors indicate forgetting.}
    \label{fig:heatmap_imbalanced}
\end{figure}

This dataset aims to investigate whether the frequency of task occurrences affects performance. Although TFCLPM achieves the highest average accuracy (cf. Table~\ref{tab:accs}), the stream analysis reveals some instability. Fig.~\ref{fig:imbalanced_result} shows the accuracy per event index, where TFCLPM initially performs well but degrades in later stages. This is further clarified by the task-specific forgetting heatmap in Fig.~\ref{fig:heatmap_imbalanced}. In these heatmaps, warm colors indicate positive knowledge retention (improvements), white indicates stability and cold colors indicate forgetting. TFCLPM shows strong retention for the first three tasks, while DynaTrainCDD \cite{kosciuszekonline} displays severe forgetting for Task 2, indicated by deep blue cells. CNAPwP demonstrates robust adaptation, maintaining positive deltas (red, light red) across task recurrences, which is an indication of effective knowledge retainment even when task frequencies vary.

\subsubsection{RecurrentTasks dataset} 
\begin{figure}[H]
    \centering
    \resizebox{\columnwidth}{!}{
\includegraphics{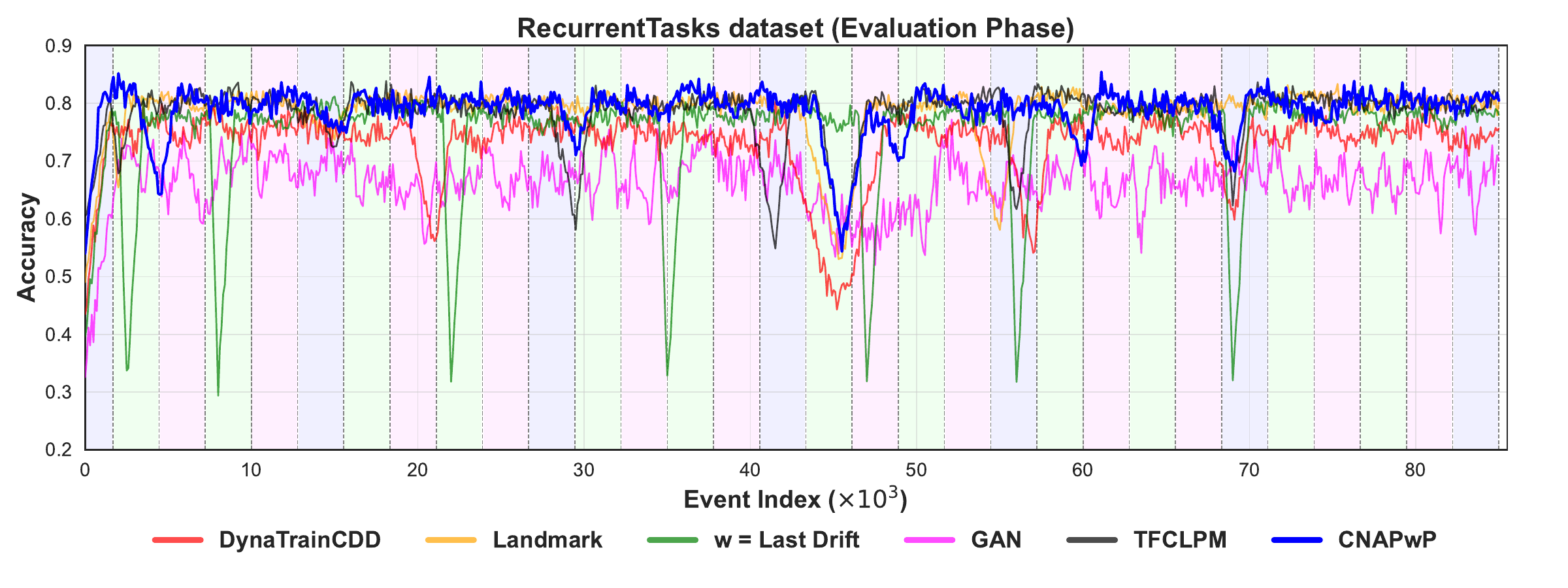}
        }
    \caption{The accuracy per event for all methods on the \texttt{RecurrentTasks} dataset. It highlights CNAPwP's ability to recover accuracy swiftly following task transitions, whereas DynaTrainCDD takes noticeably longer to regain performance. The graph also exposes the instability of the IncrementalUpdate baseline (w = LastDrift), which suffers from deep, catastrophic spikes during specific task transitions due to insufficient training data.}
    \label{fig:recurrent_result}
\end{figure}

\begin{figure}[H]
    \centering
    \resizebox{\columnwidth}{!}{
\includegraphics{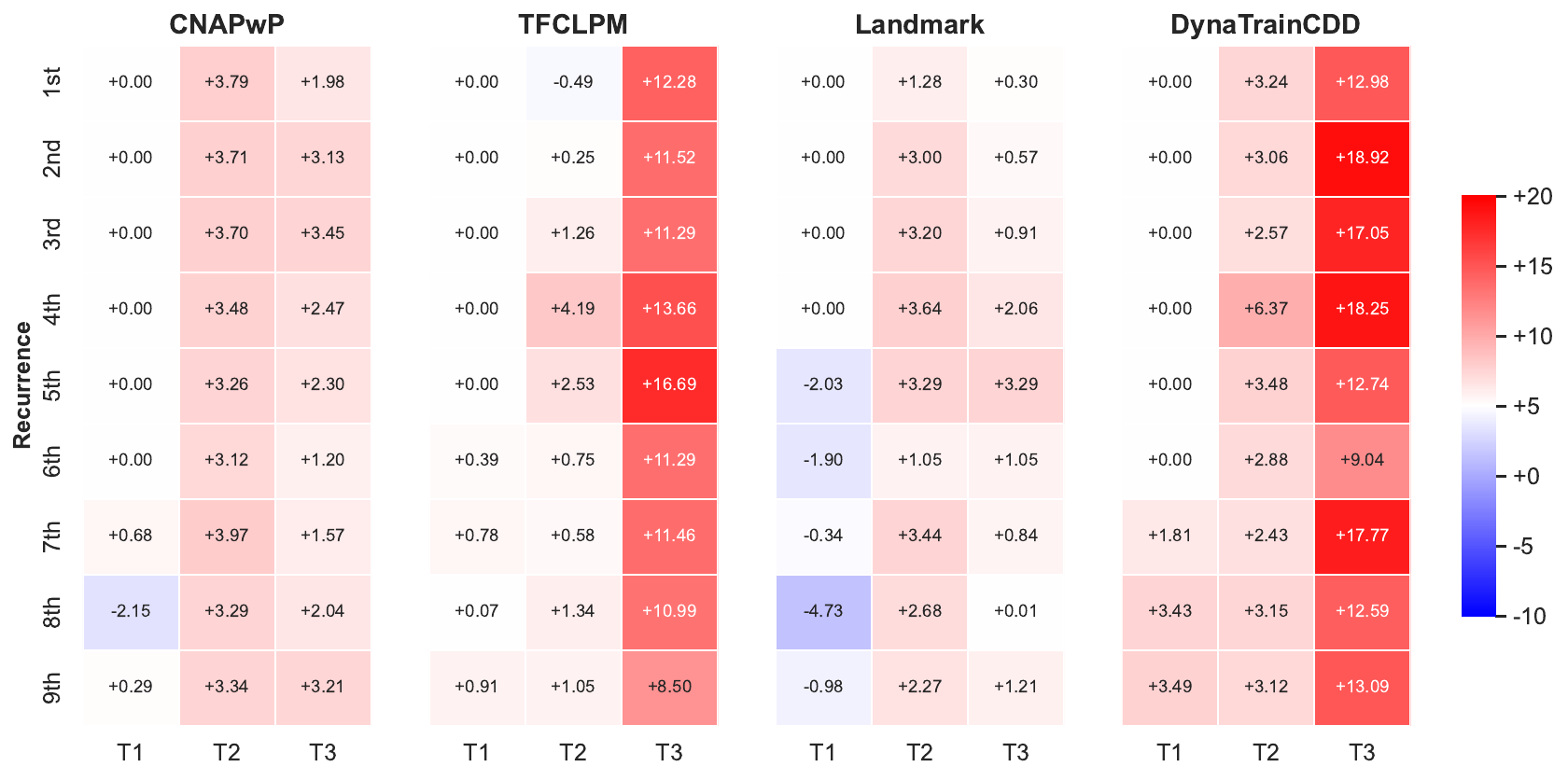}
        }
    \caption{Heatmaps showing the accuracy drop per task (task-specific forgetting) for the \texttt{RecurrentTasks} dataset for the four best-performing methods. While methods like TFCLPM and DynaTrainCDD exhibit high retention on specific tasks, CNAPwP demonstrates balanced and stable retention across all tasks, leading to the overall highest accuracy.}
    \label{fig:heatmap_recurrent}
\end{figure}

CNAPwP achieves the highest average accuracy of .789 on this dataset. Fig.~\ref{fig:recurrent_result} illustrates the accuracy per event, showing that CNAPwP recovers swiftly after task transitions compared to DynaTrainCDD, which generally takes longer to regain accuracy. The figure also highlights that while the IncrementalUpdate (w = LastDrift) \cite{pauwels2021incremental} method maintains reasonable accuracy in stable periods, it suffers from deep, catastrophic spikes during specific task transitions, indicating a lack of training data. The robustness of our approach is further analyzed in the task-specific forgetting heatmaps (cf. Fig.~\ref{fig:heatmap_recurrent}). While methods like TFCLPM and DynaTrainCDD exhibit high retention on specific tasks, CNAPwP demonstrates balanced and stable retention across all tasks. This consistent performance, avoiding the forgetting trends observed in methods like Landmark \cite{pauwels2021incremental} for Task 1, ultimately contributes to CNAPwP's superior average accuracy.

\subsubsection{RandomTasks dataset} 
\begin{figure}[H]
    \centering
    \resizebox{1.0\columnwidth}{!}{
\includegraphics{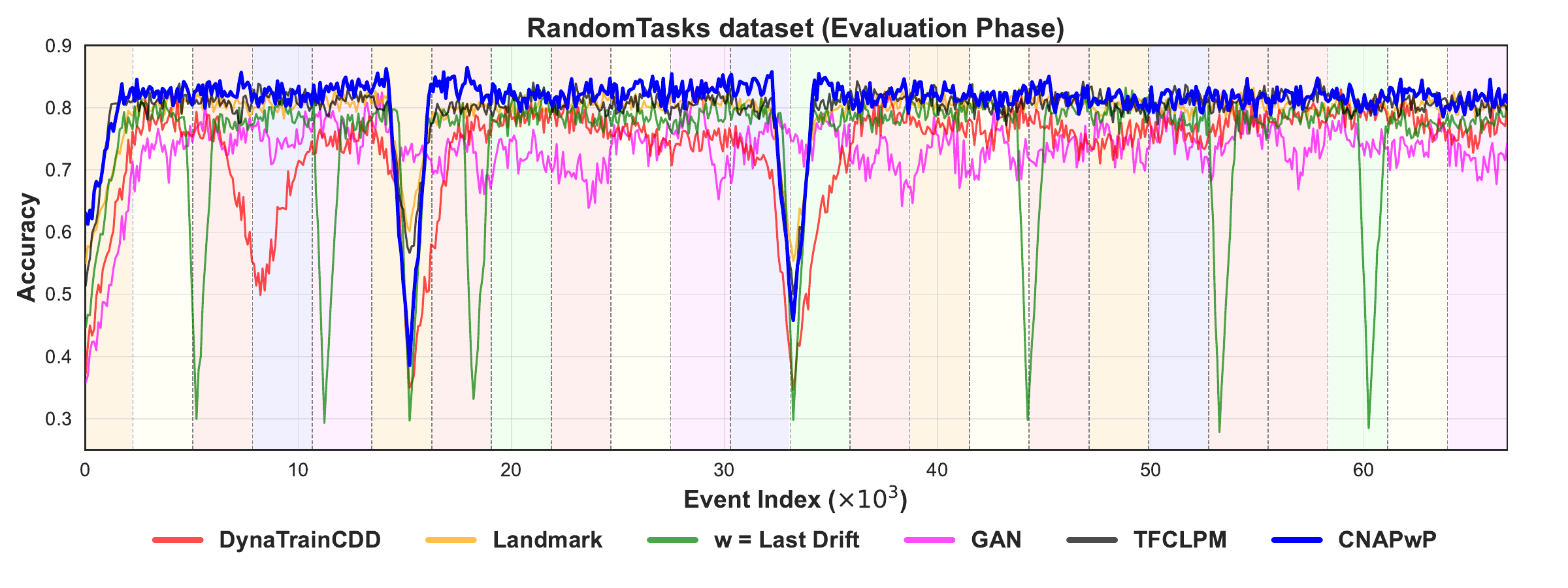}
        }
    \caption{The accuracy per event for all methods on the \texttt{RandomTasks} dataset. In this scenario of randomized task batches, CNAPwP consistently secures a higher baseline performance compared to competitors. While most methods experience significant dips during task transitions, the prompt-based approach effectively mitigates these drops in most cases, maintaining superior average accuracy across the evaluation phase.}
    \label{fig:random_result}
\end{figure}

\begin{figure}[H]
    \centering
    \resizebox{\columnwidth}{!}{
\includegraphics{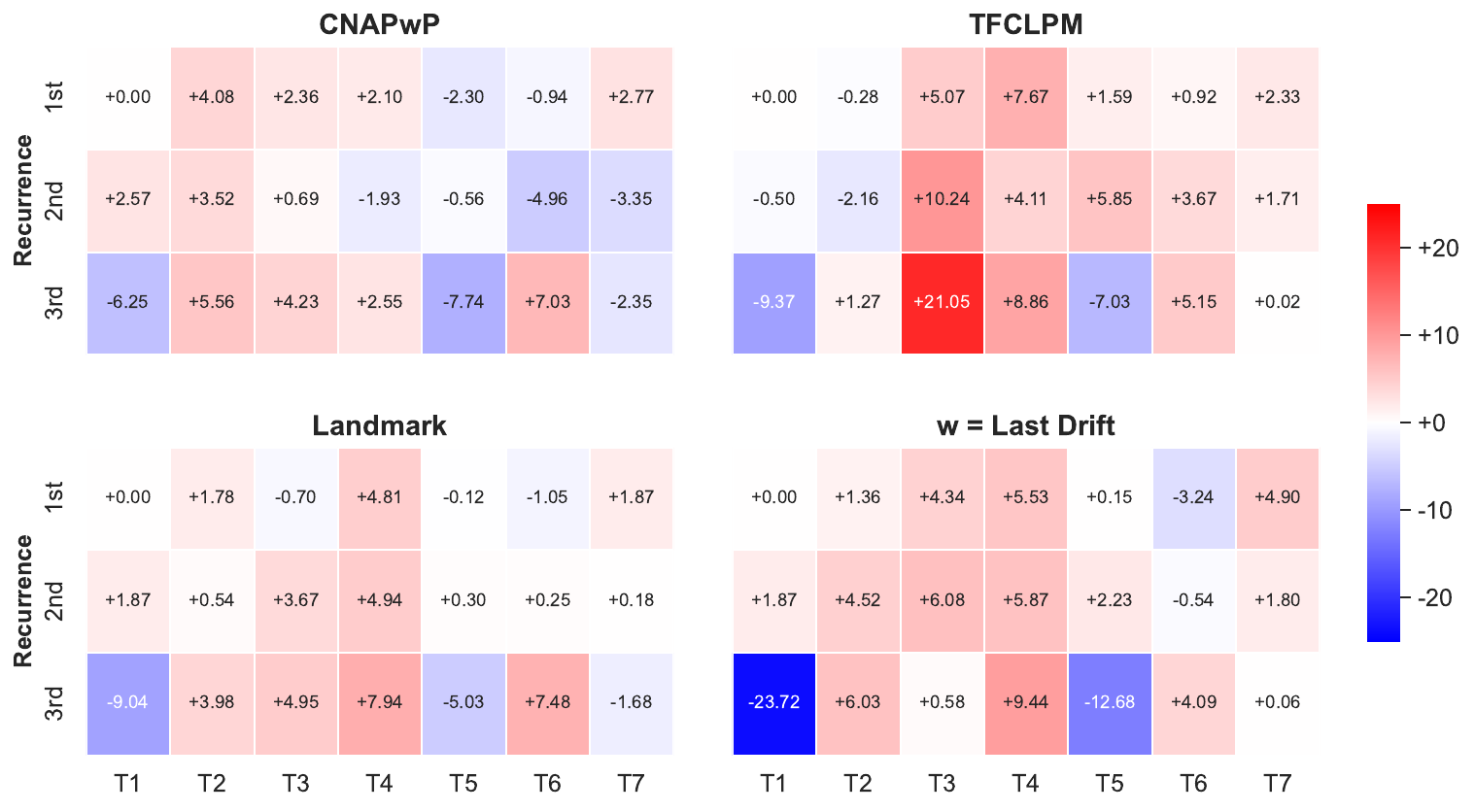}
        }
    \caption{Heatmaps showing the accuracy drop per task (task-specific forgetting) for the \texttt{RandomTasks} dataset for the four best-performing methods.The heatmaps isolate specific weaknesses in competing methods, such as the severe forgetting observed in the Incremental Update baseline on Task 1's third occurrence (-23.72\% accuracy drop). CNAPwP avoids such drastic failures, showing broader stability that outweighs the localized gains seen in methods like TFCLPM.}
    \label{fig:heatmap_random}
\end{figure}

In the RandomTasks dataset, where tasks appear in randomized batches, CNAPwP again secures the highest average accuracy. The stream accuracy in Fig.~\ref{fig:random_result} shows that while most methods dip during task transitions, CNAPwP maintains higher baseline performance. The heatmap in Fig.~\ref{fig:heatmap_random} highlights specific weaknesses of competing methods. For instance, IncrementalUpdate (w = LastDrift) suffers from severe forgetting on Task 1 in the third occurrence (-23.72). TFCLPM shows retention in Task 3 (+21.05), but the overall higher accuracy during the full stream still contributes to CNAPwP's superior accuracy.

\subsubsection{Recurrent BPIC2015 dataset} 
\begin{figure}[H]
    \centering
    \resizebox{\columnwidth}{!}{
\includegraphics{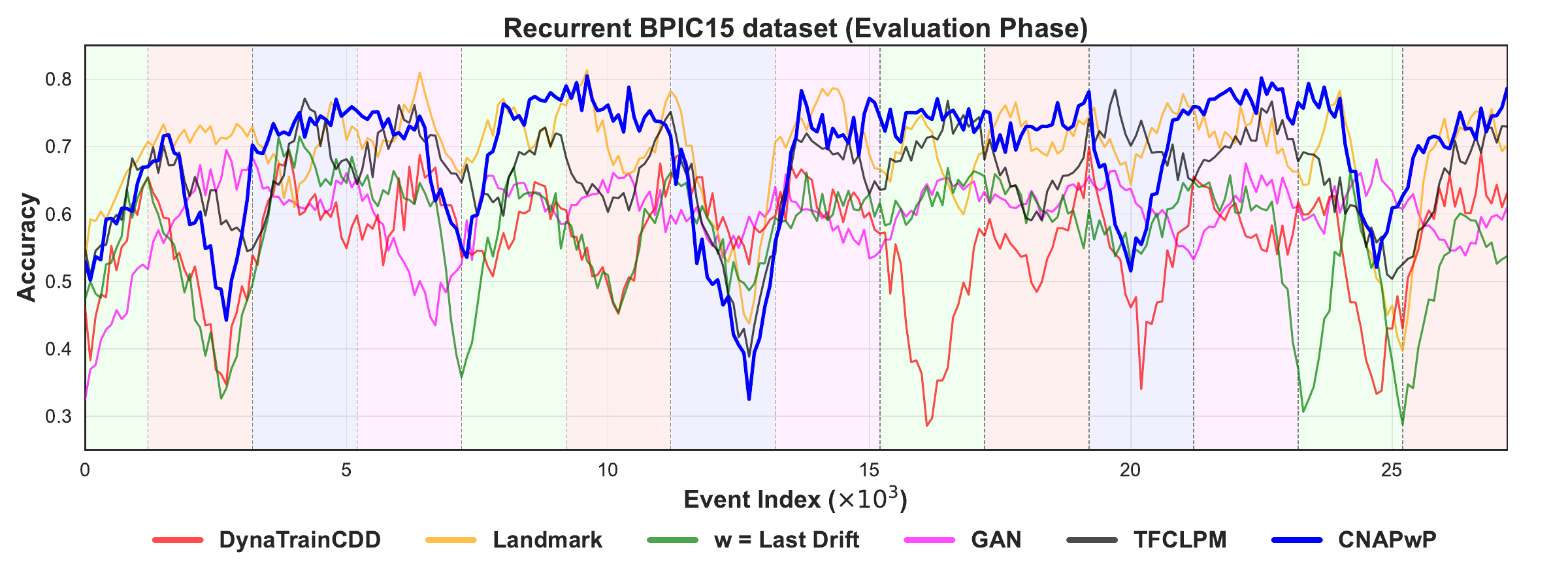}
        }
    \caption{The accuracy per event for all methods on the \texttt{Recurrent BPIC2015} dataset. The plot shows high volatility for all methods due to the complexity of the underlying process and the high number of classes. While the Landmark method achieves slightly higher overall accuracy, CNAPwP remains highly competitive and follows similar convergence patterns despite the difficult environment.}

    \label{fig:bpic15_result}
\end{figure}

\begin{figure}[H]
    \centering
    \resizebox{\columnwidth }{!}{
\includegraphics{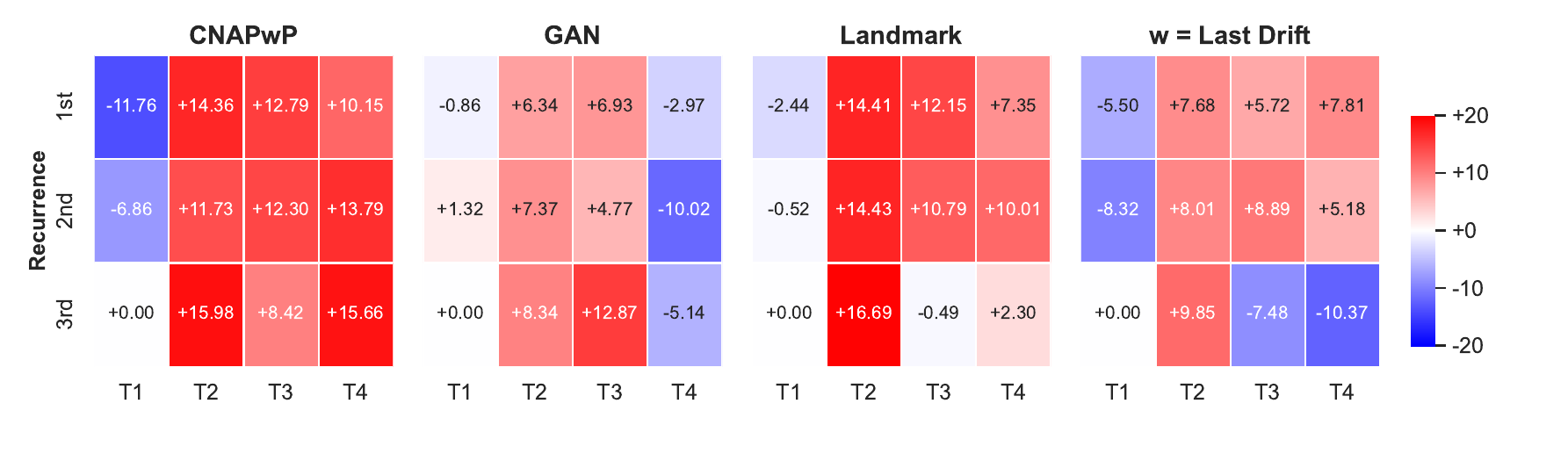}
        }
    \caption{Heatmaps showing the accuracy drop per task (task-specific forgetting) for the \texttt{Recurrent BPIC2015} dataset for the four best-performing methods. CNAPwP exhibits exceptional retention capabilities on Tasks 2, 3, and 4, indicated by deep red cells with high positive deltas where other baselines such as GAN and IncrementalUpdate (w = LastDrift). However, it does experience some localized forgetting on Task 1 in later occurrences, which slightly offsets its overall gains compared to the stable Landmark method.}
    \label{fig:bpic15_forget}
\end{figure}

This dataset presents synthetic recurrent drifts based on several different municipalities. Table~\ref{tab:accs} shows that the Landmark method achieves the highest accuracy (.670), closely followed by CNAPwP (.662) and TFCLPM (.655). The accuracy plot in Fig.~\ref{fig:bpic15_result} reveals high volatility for all methods due to both the complexity of the underlying process and the high number of classes. The task-specific forgetting heatmap (cf. Fig.~\ref{fig:bpic15_forget}) provides further insight. CNAPwP demonstrates exceptional retention capabilities on Tasks 2, 3 and 4, displaying deep red cells with high positive deltas, where other baselines often show weaker retention. However, CNAPwP suffers from significant forgetting on Task 1 in later occurrences, whereas the Landmark method remains relatively stable on this specific task. This localized degradation on Task 1 counteracts the strong gains made on other tasks, allowing Landmark to secure a slightly higher overall average accuracy.

\subsubsection{BPIC 2017 dataset} 
\begin{figure}[H]
    \centering
    \resizebox{\columnwidth }{!}{
\includegraphics{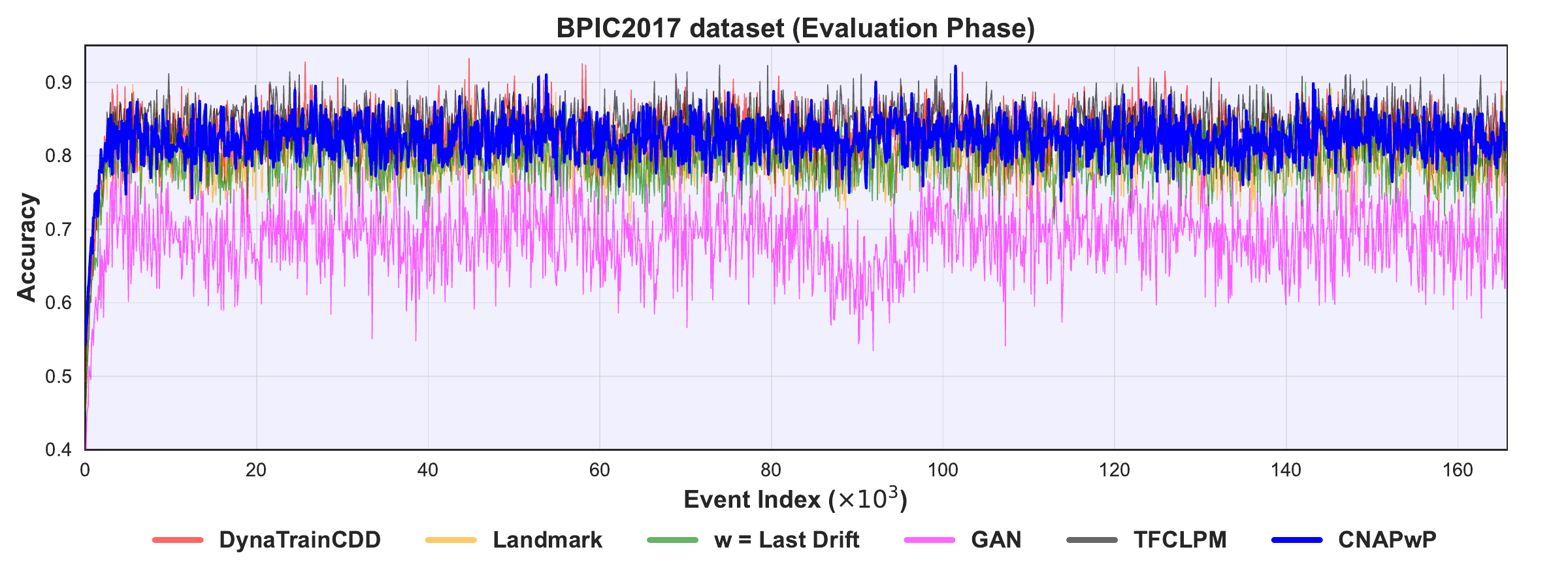}
        }
    \caption{The accuracy per event for all methods on the \texttt{BPIC 2017} dataset. This dataset (arguably) represents a stable environment with no abrupt drifts, allowing most methods to converge to high accuracy levels after a brief adaptation period. TFCLPM and CNAPwP perform similarly well, whereas the GAN-based approach struggles significantly, failing to reach the high accuracy standards set by the other methods.}
    \label{fig:bpic17_figure}
\end{figure}

Fig.~\ref{fig:bpic17_figure} displays the accuracy per event for the \texttt{BPIC2017} dataset, which (arguably) contains no drifts. Unlike the synthetic datasets with clear abrupt drifts, this real-world dataset exhibits more stable convergence patterns. TFCLPM achieves the highest average accuracy (.893) , followed by DynaTrainCDD (.820) and CNAPwP (0.818). The accuracy plot illustrates that while all methods require a brief adaptation period before converging to similar high accuracy levels, the GAN-based method \cite{taymouri2020predictive} performs significantly worse (0.693).

\paragraph{Running Time} To assess the feasibility of our approach for real-time implementation, we calculate the processing time per event, as detailed in Table~\ref{tab:time_per_event}. DynaTrainCDD consistently demonstrates the lowest running time per event across almost all datasets (ranging from 1.90 ms to 9.13 ms), benefiting from its lightweight drift-detection mechanism. In contrast, CNAPwP generally incurs a higher computational cost due to the overhead of the prefix-tuning mechanism. This difference is most pronounced in the complex \texttt{Recurrent BPIC2015} dataset, where CNAPwP requires 25.04 ms per event compared to DynaTrainCDD's 9.13 ms, which highlights how the complexity of the event log can impact our approach's processing speed.

\begin{table}[H]
    \centering
    \caption{Processing time per event and standard deviation after $5$ runs of each method over each dataset (in milliseconds).}
    \resizebox{\columnwidth}{!}{
        \begin{tabular}{lccccc}
            \toprule
            Method & \multicolumn{5}{c}{Dataset} \\
            \cmidrule(lr){2-6}
            & \texttt{RandTasks} & \texttt{ImbalTasks} & \texttt{Recur BPIC15} & \texttt{RecurTasks} & \texttt{BPIC17} \\
            \midrule
            CNAPwP (ours)                        & 3.29{\tiny$\pm$2.13}  & \underline{2.98}{\tiny$\pm$1.42}  & 25.04{\tiny$\pm$1.76}          & 3.16{\tiny$\pm$2.34}           & \underline{10.12}{\tiny$\pm$1.98} \\
            DynaTrainCDD \cite{kosciuszekonline}  & \textbf{2.06}{\tiny$\pm$1.24}  & \textbf{1.90}{\tiny$\pm$1.37}  & \textbf{9.13}{\tiny$\pm$1.84}  & \textbf{2.45}{\tiny$\pm$0.87}  & \textit{8.41}{\tiny$\pm$1.12} \\
            TFCLPM \cite{TFCLPM}                 & \textit{2.23}{\tiny$\pm$2.31}  & 4.62{\tiny$\pm$2.65}           & \textit{12.14}{\tiny$\pm$3.12} & \underline{2.81}{\tiny$\pm$1.72} & 11.02{\tiny$\pm$1.33} \\
            GAN \cite{taymouri2020predictive}     & 5.08{\tiny$\pm$2.15}  & 5.22{\tiny$\pm$2.79}           & 22.50{\tiny$\pm$1.38}          & 5.23{\tiny$\pm$1.95}           & \textbf{4.56}{\tiny$\pm$2.78} \\
            Landmark \cite{pauwels2021incremental}& 63.85{\tiny$\pm$6.47} & 18.45{\tiny$\pm$8.32}          & 229.25{\tiny$\pm$7.68}         & 33.98{\tiny$\pm$9.13}          & 13.59{\tiny$\pm$7.98} \\
            w = LastDrift \cite{pauwels2021incremental} & \underline{3.03}{\tiny$\pm$1.96} & \textit{2.41}{\tiny$\pm$1.52} & \underline{21.26}{\tiny$\pm$1.43} & \textit{2.62}{\tiny$\pm$0.93} & 13.78{\tiny$\pm$2.29} \\
            \bottomrule
        \end{tabular}
    }
    \label{tab:time_per_event}
\end{table}

However, CNAPwP remains highly competitive on \texttt{BPIC2017} (10.12 ms) , where it performs faster than both TFCLPM (11.02 ms) and IncrementalUpdate (w= LastDrift) (13.78 ms). Although CNAPwP is not consistently the fasted method, the processing time per event remains in the millisecond range (roughly 3 ms to 25 ms). Given that business process activities typically unfold over minutes or hours, this latency remains acceptable. This confirms that CNAPwP maintains near real-time performance suitable for practical deployment despite its architectural complexity.

\subsection{Ablation Study}\label{ablation}
To confirm the effectiveness of using both the E- and G-prompts, an ablation study compares the average accuracy on the \texttt{RecurrentTasks} dataset under four conditions: using only the E-prompt, using only the G-prompt, using neither prompt, and using both prompts (CNAPwP). For the baseline without prompts, a model is trained on the initial 500 events, after which its layer weights are frozen, leaving only the classification head trainable. This setup is chosen to establish a static baseline that isolates the contribution of the prompting mechanism. The value of 500 has been chosen because it represents an average window size and the GAN method also uses this value for initial training. By preventing the backbone from updating, we simulate a scenario where the model lacks the specific plasticity provided by the prompts. It allows us to quantify exactly how much knowledge retention and adaptation the G- and E- prompt contribute compared to a fixed feature extractor.

\begin{table}[H]
    \centering
    \caption{Average accuracy value for each condition on the \texttt{RecurrentTasks} dataset.}
    \resizebox{0.7\columnwidth}{!}{
        \begin{tabular}{lcccc}
            \toprule
            Condition & CNAPwP & E-Prompt & G-Prompt & No Prompt \\
            \midrule
            Average Accuracy & \textbf{.789} & .773 & .770 & .709 \\
            \bottomrule
        \end{tabular}
    }
    \label{tab:abl}
\end{table}

\begin{figure}[H]
    \centering
    \resizebox{\columnwidth}{!}{
\includegraphics{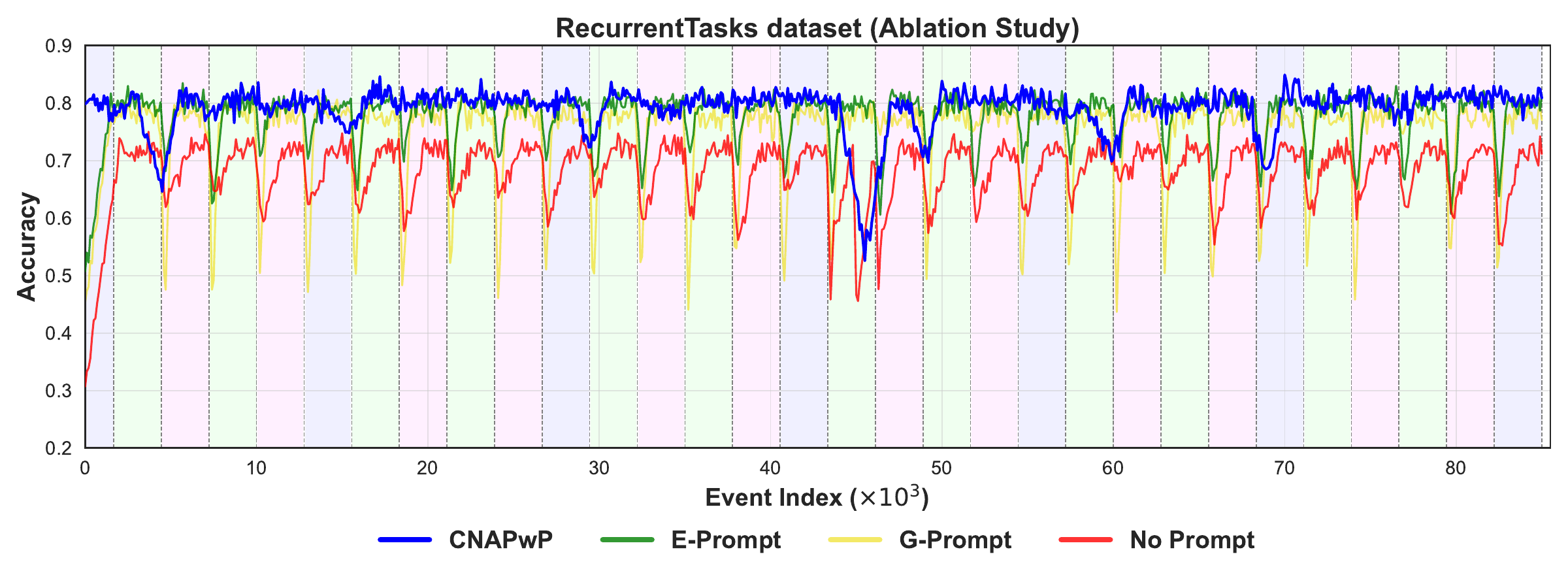}
        }
    \caption{The accuracy for each event index for the four conditions in the ablation study. It confirms that combining both E-Prompts and G-Prompts (the CNAPwP line) yields the highest performance. The distinct roles are visible: the G-Prompt helps in reducing sharp performance drops immediately after task transitions, while the E-Prompt is crucial for aiding the model's rapid recovery.}
    \label{fig:prompt_study}
\end{figure}

Table~\ref{tab:abl} reports accuracy across the four conditions. The combination of both E- and G-prompts yields the highest performance on the \texttt{RecurrentTasks} dataset, confirming their complementary benefit. Fig.~\ref{fig:prompt_study} illustrates accuracy trends by event index, revealing key differences. The G-Prompt alone reduces sharp drops after task transitions, while the E-Prompt aids recovery. Without prompts, accuracy remains consistently lower and fails to match the combined setup. These results emphasize the importance of using both prompts to enhance stability and performance.

\paragraph{Prompting Function}
The original DualPrompt study compared two prompting functions: prefix tuning and prompt tuning \cite{wang2022dualprompt}. We conducted a comparative analysis between both to determine the most effective prompting strategy for CNAPwP. As shown in Table~\ref{tab:tuning}, both approaches yield similar average accuracies across all datasets, with prefix tuning showing a slight advantage in accuracy and a significantly reduced runtime. Given the trade-off between accuracy and efficiency, prefix tuning was selected as the prompting function in CNAPwP.

\begin{table}[H]
    \centering
    \caption{Total accuracy with a prompting function, where \textbf{bold} text shows the highest accuracy or lowest running time for each prompting function and dataset.}
    \resizebox{0.9\columnwidth}{!}{
        \begin{tabular}{lcccc}
            \toprule
            & \multicolumn{2}{c}{\textbf{Prefix tuning (Pre-T)}} & \multicolumn{2}{c}{\textbf{Prompt tuning (Pro-T)}} \\
            \cmidrule(lr){2-3} \cmidrule(lr){4-5}
            & \textit{Accuracy} & \textit{Duration (s)} & \textit{Accuracy} & \textit{Duration (s)} \\
            \midrule
            \texttt{RandomTasks}        & \textbf{.813} & \textbf{264.26}  & .800 & 325.47   \\
            \texttt{RecurrentTasks}     & \textbf{.789} & \textbf{318.20}  & .780 & 401.55   \\
            \texttt{ImbalancedTasks}    & .797          & \textbf{103.82}  & \textbf{.798} & 104.08 \\
            \texttt{Recurrent BPIC2015} & \textbf{.662} & \textbf{801.80}  & .657 & 1073.63  \\
            \texttt{BPIC2017}           & \textbf{.818} & \textbf{2114.40} & .812 & 2509.76  \\
            \bottomrule
        \end{tabular}
    }
    \label{tab:tuning}
\end{table}